\documentclass[twoside,11pt]{article}

\usepackage{jmlr2e}
\usepackage{float}

\hypersetup{
    citecolor = blue
}

\usepackage{microtype}
\usepackage{graphicx}
\usepackage{subfigure}
\usepackage{booktabs} 
\usepackage[disable]{todonotes}

\usepackage{listings}
\usepackage{paralist}
\usepackage{tikz}
\usepackage{todonotes}
\usepackage{amsmath}
\usepackage{amsfonts}
\usepackage{amsopn}

\newcommand{\field}[1]{\mathbb{#1}}

\newcommand{\R}{\field{R}}

\newcommand{\vect}[1]{\boldsymbol{#1}} 
\newcommand{\mat}[1]{\boldsymbol{#1}} 
\newcommand{\tvect}[1]{\tilde{\boldsymbol{#1}}}
\newcommand{\tmat}[1]{\tilde{\boldsymbol{#1}}}
\newcommand{\tscal}[1]{\tilde{#1}}
\newcommand{\hvect}[1]{\hat{\boldsymbol{#1}}}
\newcommand{\hmat}[1]{\hat{\boldsymbol{#1}}}
\newcommand{\hscal}[1]{\hat{#1}}
\newcommand{\bvect}[1]{\bar{\boldsymbol{#1}}}
\newcommand{\bmat}[1]{\bar{\boldsymbol{#1}}}
\newcommand{\bscal}[1]{\bar{#1}}

\newcommand{\dummystring}{QWERTYU}
\newcommand{\vci}[3][\dummystr]{\ifthenelse{\equal{#1}{\dummystring}}{\vect{#2}_{#3}}{\vect{#2}_{#3}^{(#1)}}}
\newcommand{\mx}[3][\dummystr]{\ifthenelse{\equal{#1}{\dummystring}}{\mat{#2}_{#3}}{\mat{#2}_{#3}^{(#1)}}}
\newcommand{\tvci}[3][\dummystr]{\ifthenelse{\equal{#1}{\dummystring}}{\tvect{#2}_{#3}}{\tvect{#2}_{#3}^{(#1)}}}
\newcommand{\tmx}[3][\dummystr]{\ifthenelse{\equal{#1}{\dummystring}}{\tmat{#2}_{#3}}{\tmat{#2}_{#3}^{(#1)}}}
\newcommand{\tscl}[3][\dummystr]{\ifthenelse{\equal{#1}{\dummystring}}{\tscal{#2}_{#3}}{\tscal{#2}_{#3}^{(#1)}}}
\newcommand{\hvci}[3][\dummystr]{\ifthenelse{\equal{#1}{\dummystring}}{\hvect{#2}_{#3}}{\hvect{#2}_{#3}^{(#1)}}}
\newcommand{\hmx}[3][\dummystr]{\ifthenelse{\equal{#1}{\dummystring}}{\hmat{#2}_{#3}}{\hmat{#2}_{#3}^{(#1)}}}
\newcommand{\hscl}[3][\dummystr]{\ifthenelse{\equal{#1}{\dummystring}}{\hscal{#2}_{#3}}{\hscal{#2}_{#3}^{(#1)}}}
\newcommand{\bvci}[3][\dummystr]{\ifthenelse{\equal{#1}{\dummystring}}{\bvect{#2}_{#3}}{\bvect{#2}_{#3}^{(#1)}}}
\newcommand{\bmx}[3][\dummystr]{\ifthenelse{\equal{#1}{\dummystring}}{\bmat{#2}_{#3}}{\bmat{#2}_{#3}^{(#1)}}}
\newcommand{\bscl}[3][\dummystr]{\ifthenelse{\equal{#1}{\dummystring}}{\bscal{#2}_{#3}}{\bscal{#2}_{#3}^{(#1)}}}

\DeclareMathOperator{\diag}{diag}
\DeclareMathOperator*{\argmax}{arg max}

\newcommand{\abs}[1]{\left|#1\right|}

\newcommand{\eps}{\mathrm{\varepsilon}}

\renewcommand{\eqref}[1]{Eq.~\ref{eq:#1}}

\newcommand{\va}[2][\dummystring]{\vci[#1]{a}{#2}}

\newcommand{\vg}[2][\dummystring]{\vci[#1]{g}{#2}}

\newcommand{\vl}[2][\dummystring]{\vci[#1]{l}{#2}}

\newcommand{\vmu}[2][\dummystring]{\vci[#1]{\mu}{#2}}

\newcommand{\vsigma}[2][\dummystring]{\vci[#1]{\sigma}{#2}}

\newcommand{\mxf}[2][\dummystring]{\mx[#1]{F}{#2}}

\newcommand{\mxsigma}[2][\dummystring]{\mx[#1]{\Sigma}{#2}}

\newcommand{\GluonTS}{GluonTS}
\newcommand{\Swist}{GluonTS}
\usepackage[acronym]{glossaries}

\newacronym{MAPE}{MAPE}{Mean Absolute Percent Error}
\newacronym{wMAPE}{wMAPE}{Weighted Mean Absolute Percent Error}
\newacronym{MASE}{MASE}{Mean Absolute Scale Error}
\newacronym{sMAPE}{sMAPE}{Scaled Mean Absolute Percent Error}
\newacronym{CRPS}{CRPS}{Continuous Ranked Probability Score}
\newacronym{CNN}{CNN}{Convolutional Neural Network}
\newacronym{RNN}{RNN}{Recurrent Neural Network}

\makeglossaries

\newcommand{\Deepstate}{DeepState}

\def\Nc{\mathcal{N}}

\newcommand{\xit}[1]{\mathbf{x}_{i,#1}}
\newcommand{\hit}[1]{\mathbf{h}_{i,#1}}
\newcommand{\zit}[1]{z_{i,#1}}

\providecommand{\Thetait}[1]{\Theta_{i,#1}}

\newfloat{lstfloat}{htbp}{lop}
\floatname{lstfloat}{Listing}
\def\BibTeX{{\rm B\kern-.05em{\sc i\kern-.025em b}\kern-.08emT\kern-.1667em\lower.7ex\hbox{E}\kern-.125emX}}

\DeclareFixedFont{\ttb}{T1}{txtt}{bx}{n}{7} 
\DeclareFixedFont{\ttm}{T1}{txtt}{m}{n}{7}  

\usepackage{color}
\definecolor{deepblue}{rgb}{0,0,0.5}
\definecolor{deepred}{rgb}{0.6,0,0}
\definecolor{deepgreen}{rgb}{0,0.5,0}

\lstdefinestyle{numbers}{
	numbers=left,
	stepnumber=1,
	numberstyle=\tiny,
	numbersep=10pt
}
\begin{document}

\title{\GluonTS{}: Probabilistic Time Series Models in Python}

\author{\name Alexander Alexandrov%
  , 
  \name Konstantinos Benidis%
  ,
  \name Michael Bohlke-Schneider,\\
  \name Valentin Flunkert%
  ,
  \name Jan Gasthaus%
  ,
	\name Tim Januschowski,
  \name Danielle C. Maddix,
  \\
  \name Syama Rangapuram%
  ,
	\name David Salinas,
  \name Jasper Schulz,
  \name Lorenzo Stella,
  \\
  \name Ali Caner T\" urkmen%
  ,
	\name Yuyang Wang \\
	\addr Amazon Web Services
}
\editor{}

\maketitle

%
%
\begin{abstract}

We introduce Gluon Time Series (\GluonTS{})\footnote{\url{https://gluon-ts.mxnet.io}}, a library for deep-learning-based time series modeling. \GluonTS{} simplifies the development of and
experimentation with time series models for common tasks such as forecasting
or anomaly detection. It provides all necessary
components and tools that scientists need for quickly building new models, for
efficiently running and analyzing experiments and for evaluating model accuracy.
\end{abstract}

\section{Introduction}
\label{sec:intro}

Large collections of time series are ubiquitous and occur in areas as different as natural and social
sciences, internet of things applications, cloud computing, supply chains and many more. 
These datasets have substantial value, since they can be leveraged to make better forecasts or 
to detect anomalies more effectively, 
which in turn results in improved downstream decision making. Traditionally, time series modeling has
focused (mostly) on individual time series via \emph{local} models.\footnote{In local models, 
the free parameters of the time series model are estimated per individual time series in the collection of time series.} 
In recent years, advances in deep learning have led to substantial improvements over the local
approach by utilizing the large amounts of data available for estimating parameters of a single 
\emph{global}
model over the entire collection of time series. For instance,
recent publications~\citep{flunkert2017deepar,kari2017,laptev2017} and winning models of forecasting
competitions~\citep{makridakisM4concl,smyl} have shown significant accuracy 
improvements via the usage of deep learning models trained jointly on large
collections of time series.

Despite the practical importance of time series models, evidence that deep learning 
based methods lead to improved models and the success of deep-learning-based toolkits
in other domains~\citep{sockeye,mxfusion,bingham2018pyro}, there exists,
to the best of our knowledge, currently no such toolkit for time series
modeling.

We fill this gap with \Swist{} (\url{https://gluon-ts.mxnet.io}) -- a deep learning library that bundles
components, models and tools for time series applications such as forecasting or
anomaly detection. \Swist{} simplifies all aspects of scientific experiments
with time series models. It includes components such as distributions, neural 
network architectures for sequences, and feature processing steps which can be
used to quickly assemble and train new models. Apart from supporting pure
deep-learning-based models, \Swist{} also includes probabilistic models and components 
such as state-space models and Gaussian Processes. This allows scientists to combine the two
approaches, which is a promising current research direction
(e.g.,~\citep{rangapuram2018,krishnan2017structured,fraccaro2016sequential}). The
library also provides tools rapid experimentation including convenience
functions for data I/O, model evaluation and plotting utilities.
The library is based on the Apache MXNet \citep{chen2015mxnet} deep learning framework
and more specifically on the \emph{Gluon} API\footnote{\url{https://mxnet.apache.org/gluon}}.

\Swist{}'s main use-case is in building new time series models. To 
show-case its model building capabilities and facilitate 
benchmarking, it comes with 
implementations for a number of already published models. 
In this paper, we provide a
benchmark of pre-bundled models in \Swist{} on a set of public datasets.

We target \Swist{} mainly as a toolkit for scientists who are working with time
series datasets and wish to experiment with new models or solve specific
problems. Accordingly, we have designed \Swist{} such that it scales from
small to large datasets as this is the primary field of activity 
of scientists.


This paper is structured as follows. In Sec.~\ref{sec:library} we discuss the
general design principles and architecture of the library, and discuss the
different components available in \Swist{}. In Sec.~\ref{sec:problems}, we
formally introduce a number of time series problems which \Swist{} allows to address.
Sec.~\ref{sec:models} provides an overview of common neural forecasting
architectures that can be assembled with \Swist{} or are implemented as a
pre-bundled benchmark model. In Sec.~\ref{sec:experiments}, we run our benchmark
models of published deep learning based forecasting models on 11 public datasets
and demonstrate the applicability of these models to other tasks such as anomaly
detection. We discuss related work in Sec.~\ref{sec:rel_work} and conclude in
Sec.~\ref{sec:outro} with pointers to future work.

\section{Library design and components}
\label{sec:library}


In the design and structure of \Swist{}, we follow these principles.
\begin{asparadesc}
\item[Modularity:] Components are decoupled from each other and are designed
  with clear interfaces. This allows users to build models by combining and/or extending
  components in new ways.
\item[Scalability:] All components scale from running on a few time series to
  large datasets. For instance, data processing relies on streaming, so that
  the datasets do not need to be loaded into memory.
\item[Reproducibility:] To make experiments easily reproducible, 
  components serialize in a human readable way. This allows to
  ``log'' configured models and experiment setups. The user can then later fully
  reproduce or inspect the experimental setup from the log (Appendix~\ref{sec:reproducibility} contains 
  details.)
\end{asparadesc}

\autoref{lst:workflow}\footnote{All code listings in this paper are compatible with GluonTS version 0.1.3.} shows a simple workflow for creating and training a
pre-built forecasting model, and evaluating the model in a backtest. We describe the 
main components of the workflow before describing how to assemble new models.

\begin{lstfloat}[t]
\begin{lstlisting}[language=python,style=numbers,emph={DatasetRepository,DeepAREstimator,Evaluator,Trainer,MyTrainNetwork,MyPredNetwork,GluonForecastEstimator,HybridBlock,QuantileForecast,InstanceSplitter,__init__}, % Custom highlighting
]
from gluonts.dataset.repository.datasets import get_dataset
from gluonts.model.deepar import DeepAREstimator
from gluonts.trainer import Trainer
from gluonts.evaluation import Evaluator
from gluonts.evaluation.backtest import backtest_metrics

meta, train_ds, test_ds = get_dataset("electricity") 

estimator = DeepAREstimator(|\label{lst:estimator}|
freq=meta.time_granularity,
prediction_length=100,
trainer=Trainer(epochs=20, batch_size=32)
)

predictor = estimator.train(train_ds)|\label{lst:train}|

evaluator = Evaluator(quantiles=(0.1, 0.5, 0.9))
agg_metrics, item_metrics = backtest_metrics(
train_dataset=train_ds,
test_dataset=test_ds,
forecaster=predictor,
evaluator=evaluator
)

\end{lstlisting}
\caption{Model training and evaluation in \Swist{}}
\label{lst:workflow}
\end{lstfloat}

\subsection{Data I/O and processing}

\Swist{} has two types of data sources that allow a user to experiment and
benchmark algorithms. The first is a \lstinline|DatasetRepository| that contains
a number of public time series datasets, and is extensible with custom private
datasets. These input dataset can be included in jsonlines or parquet format.
\todo[inline]{Tim: given that we've taken this out of the open-sourcing, we may 
want to revisit the DatasetRepository.}

The second data source is a generator for synthetic time series. The
user can configure the generator in a declarative way to generate complex time
series that exhibit e.g.\ seasonality, trend and different types of noise. Along
with the time series itself, the generator can produce different types of
co-variates, such as categorical co-variates, and allow the properties of the
generated time series to depend on these. 

From a high level perspective, data handling in \Swist{} is done on streams
(Python iterators) of dictionaries. During feature processing, a
\lstinline|DatasetReader| loops over the input dataset, and emits a stream of
dictionaries. The feature processing pipelines consists of a sequence of
\lstinline|Transformation|s that act successively on the stream of dictionaries.
Each \lstinline|Transformation| can add new keys or remove keys, modify values,
filter items from the stream or add new items.

\Swist{} contains a set of time series specific transformations that include 
splitting and padding of time series (e.g.\
for evaluation splits), common time series transformation such as Box-Cox
transformations or marking of special points in time and missing values.
A user can easily include custom transformations for specific purposes,
and combine them with existing transformations in a pipeline.

\subsection{Predictor}

\Swist{} follows a stateless predictor API as is e.g.\ used in
scikit-learn~\citep{scikitlearn} and SparkML~\citep{sparkml}. An
\lstinline|Estimator| has a \lstinline|train| method that takes a (train)
dataset and returns a \lstinline|Predictor|. The \lstinline|Predictor| can then
be invoked on a (test) dataset or a single item to generate predictions.
Classical forecasting models that estimate parameters per time series before
generating predictions are pure predictors in this API. \Swist{} faciliates 
comparisons with established forecasting packages (e.g.~\citep{hyndman2008R,prophet17}).

\subsection{Distribution / Output}
\label{sec:distribution}

\Swist{} contains a flexible abstraction for probability distributions (and densities), 
which are common building blocks in
probabilistic time series modeling. Concrete implementations include common parametric distributions, such
as Gaussian, Student's $t$, gamma, and negative binomial. There is also a
binned distribution that can be used to quantize a signal. The binned
distribution is configured with bin edges and bin values that represent the
bins, e.g.\ when samples are drawn.\todo[inline]{Tim: How about replacing the sentence before with this:
This distribution is parametrized via the boundaries and the representative 
value of each bin.
}

A \lstinline|TransformedDistribution| allows the user to map a distribution using
a differentiable bijection~\citep{tensorflow_distributions, rezendeVariationalInferenceNormalizing2015}. 
A bijection can be a simple fixed
transformation, such as a log-transform that maps values into a different domain. 
\todo[inline]{Tim: strictly speaking, the logarithm isn't bijective. Later we talk 
about "invertible" instead. Maybe that's the better term?}
 Transformations can also be arbitrarily complex (as long as they are
invertible, and have a tractable Jacobian determinant), and can depend
on learnable parameters that will be estimated jointly with the other model
parameters.  In particular, this paradigm covers the popular Box-Cox
transformations~\citep{boxcox1964}. \Swist{} also supports mixtures of
arbitrary base distributions. \todo[inline]{Tim: the sentence about 
mixtures of base distributions, should we move this upwards to the parametric 
distributions?}

\subsection{Forecast object}

Trained forecast models (i.e.\ predictors) return \lstinline|Forecast| objects
as predictions, which contain a time index (start, end and time granularity) and
a representation of the probability distribution of values over the time index.
Different models may use different techniques for estimating and representing
this joint probility distribution.

The auto-regressive models (Section~\ref{sec:models:auto_reg}) 
naturally generate sampled trajectories, and in this
case, the joint distribution is represented through a collection of (e.g.
$n=1000$) sample paths -- potential future times series trajectories. Any desired
statistic can then be extracted from this set of sample paths. For instance, in
retail, a distribution of the total number of sales in a time interval can be
estimated, by summing each individual sample path over the interval. The result
is a set of samples from the distribution of total sales~\citep{seeger2016,flunkert2017deepar}.

Other models output forecasts in the form of quantile predictions for a set of 
quantiles and time intervals \citep{kari2017, sqf}. The
corresponding \lstinline|QuantileForecast| then stores this collection of
quantile values and time intervals.

Forecast objects in \Swist{} have a common interface that allows
the \lstinline|Evaluation| component to compute accuracy metrics. In
particular, each forecast object has a
\lstinline|.quantile| method that takes a desired percentile value (e.g. $0.9$) and
an optional time index, and returns the estimate of the corresponding quantile
value in that time index.

\subsection{Evaluation}
\label{sec:evaluation}
For quantitatively assessing the accuracy of its time series models, 
\Swist{} has an \lstinline|Evaluator| class that can be
invoked on a stream of \lstinline|Forecast| objects and a stream of true
targets (pandas data frame).

\begin{figure}
  \centering
  \includegraphics[width=.6\columnwidth]{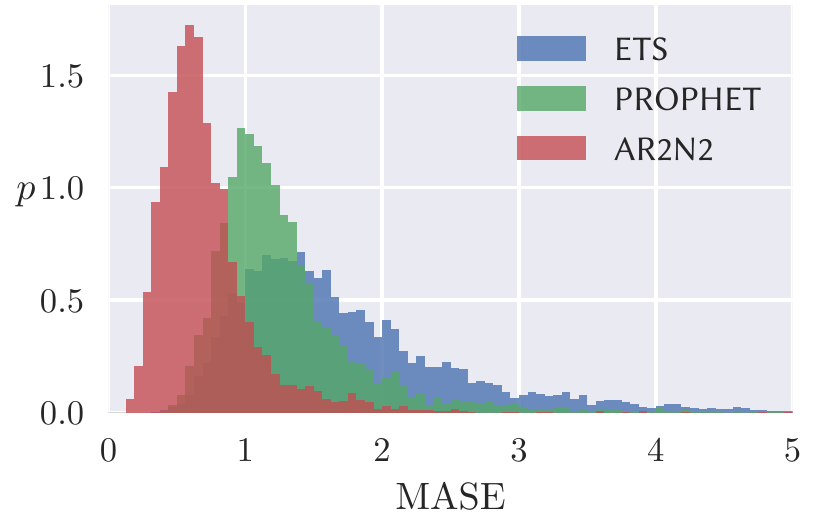}
  \caption{Histogram of forecast error measured in \gls{MASE} on traffic-dataset for ETS, Prophet
    and DeepAR.}
 \label{fig:hist_mase}
\end{figure}

The \lstinline|Evaluator| iterates over true targets and forecasts in a
streaming fashion, and calculates metrics, such as quantile loss, \gls{MASE},
\gls{MAPE} and \gls{sMAPE}. The \lstinline|Evaluator| return these per item
metrics as a pandas data frame. This makes it easy for a user to explore or
visualize statistics such as the error distribution across a data set.
Figure~\ref{fig:hist_mase} shows an example histogram of \gls{MASE} values for
the ETS, Prophet and DeepAR method (see Sec.~\ref{sec:models} for details on these models)
on the same dataset. In addition to the per item metrics, the
\lstinline|Evaluator| also calculates aggregate statistics over the entire
dataset, such as \gls{wMAPE} or weighted quantile loss, which are useful for instance for
automatic model tuning e.g.\ using SageMaker.

For qualitatively assessing the accuracy of time series models, \Swist{}
contains methods that visualize time series and forecasts using
matplotlib~\citep{matplotlib}.

In addition to the \lstinline|Evaluator|, \Swist{} allows the user to run full
backtest evaluations. The \lstinline|backtest_metrics| method runs a test
scenario that trains an estimator on a training dataset, or uses a pre-trained
predictor, and evaluates the model on a test dataset. A separate
\lstinline|Splitter| component generates train/validation/test splits from an
existing dataset. These splits can be simple single time point splits or more
complex test sets supporting for instance evaluation over multiple different
time points or rolling windows. A user can easily combine these components to
run simple or complex training and evaluation scenarios in a reproducible manner.

%
%

\subsection{Time Series Model Building}
\label{sec:modelbuilding}

\Swist{}'s primary purpose is the creation of new models. A user can implement a new model
fully from scratch by just adhering to the
\lstinline|Estimator| / \lstinline|Predictor| API. In most cases however, it is
simpler to use a pre-structured model template. For instance, to create a new deep
learning based forecast model, one can use the
\lstinline|GluonForecastEstimator| and implement and configure
the following members and methods:
\begin{itemize}
\item \lstinline|train_network_cls|: A neural network that returns the training
  loss for the model and will be called during training.
\item \lstinline|pred_network_cls|: A neural network that uses the same parameters as
  the training network, but instead of returning a loss returns the forecast samples.
\item \lstinline|create_transformation|: A method that returns the
  transformation pipeline.\footnote{Since the transformation pipeline can itself contain
  learnable parameters and the \lstinline|Estimator| is stateless this is a
  method returning the transformation rather than a fixed member.}
\item \lstinline|forecast_cls|: The forecast class that will be wrapped around
  the output of the forecast network.
\end{itemize}
Appendix~\ref{sec:newmodel} presents an example containing a simple 
model that estimates the median for each future time point independently.\todo[inline]{Tim: maybe 
replace the example in the appendix with a more complex example where we use distributions or even mixtures?} We describe 
more complex models in Sec.~\ref{sec:models} which can be implemented readily in \Swist{}
 with the abstractions shown above.

\section{Time Series Problems}\label{sec:problems}

\Swist{} addresses probabilistic modeling of uni- or multi-variate sequences of
(large) collections of time series. Important applications include 
forecasting, smoothing and anomaly detection.

More formally,
let $Z= \{\zit{1:T_i}\}_{i=1}^N$ be a set of $N$ univariate time series, where
$\zit{1:T_i} := (\zit{1}, \zit{2}, \ldots, \zit{T_i})$, and $\zit{t} \in \mathbb{R}$ denotes the value of the $i$-th time series at time $t$. We mainly consider time series where the time points are equally spaced (Sec.~\ref{sec:models:pps} is an exception) but the time units across different sets can be arbitrary (e.g. hours, days, months).
  Further, the time series do not have to be aligned, i.e., the starting point $t=1$ can refer to a different absolute time point for different time series $i$.

Furthermore, let $X= \{\xit{1:T_i+\tau}\}_{i=1}^N$ be a set of associated, 
time-varying covariate vectors with $\xit{t} \in \mathbb{R}^{D}$.

The goal of \emph{forecasting}~\citep{hyndman2017forecasting} is to predict the probability 
distribution of future values $\zit{T_{i+1}:T_{i + \tau}}$ given the past values $\zit{1:T_i}$, 
the covariates $\xit{1:T_{i+\tau}}$, and the model parameters $\Phi$:
\begin{equation}
p\left(\zit{T_i + 1:T_i+\tau}\ |\ \zit{1:T_i}, \xit{1:T_i+\tau}; \Phi \right).
\label{eq:forecast_dist}
\end{equation}

\emph{Smoothing} or \emph{missing value imputation} in time series can 
leverage the sequence structure, and therefore allow for more sophisticated 
approaches compared to the general missing value imputation setting (e.g.,~\citep{missingValue}). 
Missing values are 
commonly found in real-world time series collections. For instance, in
retail demand forecasting, the true demand is not observed when the item
is not in-stock, i.e., cannot be sold.
Smoothing is similar to forecasting, except that the time points that we want to predict do not lie in
the future. Instead, for a set of series and \emph{arbitrary} time points there are missing or unobserved values, and the goal is to estimate the (joint) probability
distribution over these missing values, i.e.,
\begin{equation*}
p\left(\zit{j_1:j_\tau}\ |\ \zit{k_1:k_\tau}, \xit{k_1:x_{k\tau}}; \Phi \right).
\end{equation*}
where $\{j_1:j_\tau\}$ are the indexes of the  missing values 
and $\{k_1:k_\tau\}$ the indexes of the observed values, such that $\{j_1:j_\tau\} \cup \{ k_1:k_\tau \} = \{1:T_i\}$.

In \emph{anomaly} or \emph{outlier detection} we want to identify time points,
where the observed values are unlikely to have occurred. While we may have
additional labels that annotate anomalous behavior, it is in many applications
not feasible to directly train a classifier on these labels, because the labels
are too sparse -- after all, anomalies are often rare.

In this unsupervised case, anomaly detection is similar to forecasting, except
that all values are observed and we want to know how likely they were. A
probabilistic forecasting model can be converted into an anomaly detection model
in different ways. For univariate models where the cumulative distribution
function (CDF) of the marginal predicted distribution can be evaluated, one can
directly calculate the $p$-value (tail probability) of a newly observed data
point via the CDF \citep{shipmonTimeSeriesAnomaly2017}. For other models or in
the multivariate case, we can use the log-likelihood values. This is slightly
more complicated, since the range of log-likelihood values depends on the model
and the dataset, however, we can estimate multiple high percentiles of the
distribution of negative log-likelihood values on the training set, such as the
99, 99.9, 99.99 etc. At prediction time, we can evaluate the negative
log-likelihood of each observations under the model sequentially, and compare
this with the estimated quantiles. Both approaches allow us to mark unlikely
points or unlikely sequences of observations as anomalies. The quality of this
forecasting based approach depends on the frequency of such anomalies in the
training data, and works best if known anomalies are removed or masked. If some
labels for anomalous points are available, they can be used to set likelihood
thresholds or to evaluate model accuracy e.g.\ by calculating classification
scores, such as precision and recall.

\section{Time Series Models}\label{sec:models}

All tasks from Sec.~\ref{sec:problems} have at their core the estimation of a
joint probability distribution over time series values at different time points.
We model this as a supervised learning problem, by fixing a model structure
upfront and learning the model parameters $\Phi$ using a statistical
optimization method, such as maximum likelihood estimation, and the sets $Z$ and
$X$ as training data.

Classical models that were developed for these tasks
\citep{hyndman2017forecasting}, are, with some exceptions \citep{chapados2014},
local models that learn the parameters $\Phi$ for each time
series individually. Recently, however, several neural time series models have
been proposed~\citep{flunkert2017deepar,sqf,rangapuram2018,laptev2017,kari2017}
where a single global model is learned for all time series in the dataset by
sharing the parameters $\Phi$.\footnote{While neural network based forecasting
methods have been used for a long time, they have been traditionally employed
as local models. This led to mixed results compared with other local modeling
approaches~\citep{zhang1998forecasting}.} Due to their capability to extract
higher-order features, these neural methods can identify complex patterns within
and across time series from datasets consisting of raw time
series~\citep{flunkert2017deepar,laptev2017} with considerably less human effort.

Time series models can be broadly categorized as generative and discriminative, 
depending on how the target $Z$ is modeled \citep{ng02}.\footnote{Note that our 
categorization overloads the distinction used in Machine Learning.
Technically, neither of the category defined here jointly models the covariates 
$X$ and the target $Z$ and thus both belong to ``discriminative'' models in the traditional sense.} 
\begin{center}
\begin{tabular}{c|l}
\toprule
\textsc{category} & \textsc{modeling}\\
\midrule
generative & $p(\zit{1:T_i+\tau}|\xit{1:T_i+\tau}; \Phi)$ \\
discriminative & $p(\zit{T_i + 1:T_i+\tau}\ |\ \zit{1:T_i}, \xit{1:T_i+\tau}; \Phi)$\\
\bottomrule
\end{tabular}
\end{center}
Generative models assume that the given time series are generated from an unknown stochastic process $p(Z | X; \Phi)$ given the covariates $X$.
The process is typically assumed to have some parametric structure with unknown parameters $\Phi$.
Prominent examples include classical models such as ARIMA and ETS~\citep{hyndman2008}, Bayesian structural time series (BSTS)~\citep{Scott2014PredictingTP} and the recently proposed deep state space model ~\citep{rangapuram2018}. 
The unknown parameters of this stochastic process are typically estimated by maximizing the likelihood, which is the probability of the observed time series, $\{\zit{1:T_i}\}$, under the model $p(Z | X; \Phi)$, given the covariates $\{\xit{1:T_i}\}$.
Once the parameters $\Phi$ are learned, the forecast distribution in Eq.~(\ref{eq:forecast_dist}) can be obtained from $p(Z | X; \Phi)$.
In contrast to ETS and ARIMA, which learn $\Phi$ per time series individually, neural generative models like~\citep{rangapuram2018} further express $\Phi$ as a function of a neural network whose weights are shared among all time series and learned from the whole training data.


Discriminative models such as~\citep{flunkert2017deepar,sqf,kari2017}, model the conditional distribution (for a fixed $\tau$) from Eq.~(\ref{eq:forecast_dist}) directly via a neural network. 
Compared to generative models, conditional models are more flexible, since they make less structural assumptions, and hence are also applicable to a broader class of application domains. 
In the following, we distinguish between auto-regressive and sequence-to-sequence models among discriminative models. 
Further distinctions within the subspace of discriminative models are discussed in more detail in~\citep{Faloutsos2018}.

\subsection{Generative Models}
Here we describe in detail the forecasting methods implemented in \Swist{} that fall under the generative models category.
Further examples in this family include Gaussian Processes~\citep{girard2003gaussian} and 
Deep Factor models~\citep{maddix2018deep,wang2019deep} which we omit due to space restrictions.
%
%
\textbf{State Space Models} (SSMs) provide a principled framework for modeling and learning time series patterns~
\citep{hyndman2008,durbin2012time,seeger2016}.
In particular, SSMs model the temporal structure of the data via a latent state $\vl{t} \in \R^{D}$ that can be used to
 encode time series components, such as level, trend, and seasonality patterns.
A general SSM is described by the so-called state-transition equation, defining the stochastic transition 
dynamics $p(\vl{t}|\vl{t-1})$ by which the latent state evolves over time, and an observation model specifying the 
conditional probability $p(z_t|\vl{t})$ of observations given the latent state.
Several classical methods (e.g., ETS, ARIMA) can be cast under this framework by choosing appropriate 
transition dynamics and observation model~\citep{hyndman2008}.

A widely used special case is the linear innovation state space model, where the transition dynamics and the observation model are given by. Note that we drop the index $i$ from the notation here since these models are typically applied to individual time series.
\begin{align*}
    \vl{t} &= \mxf{t}\vl{t-1} + \vg{t}\eps_t, \quad && \eps_t\sim \Nc(0, 1),\\
    z_t &= y_t + \sigma_t \epsilon_t, \quad y_t = \va{t}^{\top}\vl{t-1} + b_t,\quad &&\epsilon_t \sim \mathcal{N}(0, 1).
\end{align*}
Here $\va{t} \in \mathbb{R}^L$, $\sigma_t\in \mathbb{R}_{>0}$ and $b_t \in \mathbb{R}$ are further (time-varying) parameters of the model.
Finally, the initial state $\vl{0}$ is assumed to follow an isotropic Gaussian distribution, \hbox{$\vl{0} \sim N(\vmu{0}, \diag(\vsigma{0}^2))$}.

The state space model is fully specified by the parameters $\Theta_t = (\vmu{0}, \mxsigma{0}, \mxf{t}, \vg{t}, \va{t}, b_t, \sigma_t)$, $\forall t>0$.
One generic way of estimating them is by maximizing the marginal likelihood, i.e., 
\[ \Theta^*_{1:T} = \argmax_{\Theta_{1:T}} p_{SS}(z_{1:T} | \Theta_{1:T}),\]
where $p_{SS}(z_{1:T}| \Theta_{1:T})$
denotes the marginal probability of the observations $z_{1:T}$ given the parameters $\Theta$ under the state space model, integrating out the latent state $\vl{t}$. 

\textbf{Deep State Space Models}~\citep{rangapuram2018} (referred as \Deepstate\ here) is a probabilistic time series forecasting approach that combines state space models with deep learning.
The main idea is to parametrize the linear SSM using a recurrent neural network (RNN) whose weights are learned jointly from a dataset of raw time series and associated covariates.
More precisely, \Deepstate\ learns a globally shared mapping $\Psi(\xit{1:t}, \Phi)$ from the covariate vectors $\xit{1:T_i}$ associated with each target time series $\zit{1:T_i}$, to the (time-varying) parameters $\Thetait{t}$ of a linear state space model for the $i$-th time series.
The mapping $\Psi$ from covariates to state space model parameters is parametrized using a deep recurrent neural network (RNN).
Given the covariates\footnote{The covariates (features) can be time dependent (e.g. product price or a set of dummy variables indicating day-of-week) or time independent (e.g., product brand, category etc.).} $\xit{t}$ associated with time series $\zit{t}$, %
a multi-layer recurrent neural network with LSTM cells and parameters $\Phi$, it computes a representation of the features via a recurrent function $h$,
\[  \hit{t} = h(\hit{t-1}, \xit{t}, \Phi). \]
The real-valued output vector of the last LSTM layer is then mapped to the parameters $\Thetait{t}$ of the state space model, by applying affine mappings followed by suitable elementwise transformations constraining the parameters to appropriate ranges.

Then, given the features $\xit{1:T}$ and the parameters $\Phi$, under this model, the data $\zit{1:T_i}$ is distributed according to
\begin{equation}
  p(\zit{1:T_i} | \xit{1:T_i}, \Phi) = p_{SS} (\zit{1:T_i} | \Thetait{1:T_i}), \quad \quad  i= 1, \ldots, N.
\end{equation}
where $p_{SS}$ denotes the marginal likelihood under a linear state space model, given its (time-varying) parameters $\Thetait{t}$.
Parameters $\Thetait{t}$ are then used to compute the likelihood of the given observations $\zit{t}$, which is used for learning of the network parameters $\Phi$.

\subsection{Discriminative Models}
Inspired by the sequence-to-sequence learning approach presented in~\citep{sutskever2014}, several forecasting methods are proposed in this framework~\citep{kari2017}.
These sequence-to-sequence models consist of a so-called
encoder network that reads in a certain context of the training range of the time series (i.e., $\zit{1:T_i}, \xit{1:T_i}$), and 
encodes information about the sequence in a latent state.
This information is then passed to the decoder network, which generates the forecast $\zit{T_i + 1:T_i+\tau}$ by combining the latent information with the features $\xit{T_i + 1 : T_i + \tau}$ in the prediction range. 

While sequence-to-sequence models are more complex and may need more training
data compared to simple auto-regressive models, they have several advantages.
First, they can handle differences in the set of features available in the
training and prediction ranges. This enables the use of time series covariates
that are not available in the future, which is a typical problem often
encountered in practical applications. Second, the decoder does not have to be
an auto-regressive architecture. During the multi-step ahead prediction, the
prediction from the previous step does not need to be used as a feature for the
current time step. This avoids error accumulation, which can be beneficial
especially for long forecast horizons.

Compared to auto-regressive models, which represent the sequential nature of the
time series well, a sequence-to-sequence approach is more akin to multivariate
regression. This means that the prediction horizon $\tau$ has to be fixed
beforehand in sequence-to-sequence models, and a complete retraining is needed
if the forecast is required beyond $\tau$ steps.

\Swist{} contains a flexible sequence-to-sequence framework that makes it
possible to combine generic encoder and decoder networks to create custom sequence-to-sequence models. 
Moreover, \Swist{} also has example implementations of specific forecasting models~\citep{kari2017} that fall under this category as well as generic models~\citep{attention} that have been proven to be successful in this domain.

\textbf{Neural Quantile Regression Models.} 
Quantile regression \citep{koenker_2005} is a technique for directly predicting a particular quantile of a dependent variable.
These techniques have been combined with deep learning and employed in the context of time series forecasting \citep{Xu2016quantile, kari2017}.
More specifically, within the sequence-to-sequence framework, we can employ a decoder that directly outputs a set of quantile values for each time step in the forecast horizon, and 
train such a model using the quantile loss function.


We have implemented variants of such quantile decoder models in \Swist{} following~\citep{kari2017} and combined them with RNN and Dilated Causal Convolution (CNN) encoders, resulting in models dubbed RNN-QR and CNN-QR, respectively.


\textbf{Transformer.}
\Swist{} also contains an implementation of the Transformer architecture
\citep{attention} that has been successful in natural language processing. The Transformer model captures the dependencies of a sequence by relying entirely on attention mechanisms.  The elimination of the sequential
computation makes the representation of each time step independent of all other time steps and therefore allows the parallelization of the computation.

In a nutshell, the Transformer architecture uses stacked self-attention and point-wise, fully connected layers for both the encoder and the decoder, while the decoder has an additional cross-attention layer that
is applied on the encoder output. In our implementation we use the same
architecture as in \citep{attention}. However, since \citep{attention} focuses on
the natural language processing problem, we replace the softmax distribution over discrete values (e.g.\ characters or words) by the generic distribution component discussed in Sec.~\ref{sec:distribution} so our model fits the probabilistic time series forecasting framework.\textbf{}

\subsection{Auto-regressive models}\label{sec:models:auto_reg}
Auto-regressive models reduce the sequence prediction task fully to a
one-step-ahead problem where we model \[p\left(\zit{T_i + 1} |\ \zit{1:T_i}, \xit{1:T_i+\tau}; \Phi \right).\]
 The model is trained on a
sequence by sequential one-step-ahead predictions, and the error is aggregated
over the sequence for the model update. For prediction, the model is propagated
forward in time by feeding in samples,  Through multiple simulations, a set of sample-paths representing the joint probability distribution
over the future of the sequence is obtained.

\textbf{NPTS.} The Non-Parametric Time Series forecaster (NPTS)~\citep{gasthaus2016} falls into the
class of simple forecasters that use one of the past observed targets as the
forecast for the current time step.
Unlike the naive or seasonal naive forecasters that use a fixed time index (the previous index $T-1$ or the past season $T-\tau$) as the prediction for the time step $T$, NPTS randomly samples a time index $t \in \{0, \ldots, T-1\}$ in the past to generate a prediction sample for the current time step $T$. By sampling multiple times, one obtains a Monte Carlo sample from the predictive distribution, which can be used e.g.\ to compute prediction intervals.
More precisely, given a time series $z_0, z_1, \ldots, z_{T-1}$, the generative process of NPTS for time step $T$ is given by 
\[   
    \hat{z}_T = z_t, \quad t \sim q_T(t), \quad  t \in \{0, \ldots, T-1\}, 
\]
where $q_T(t)$ is categorical probability distribution over the indices in the training range.
Note that the expectation of the prediction, $\mathbb{E}[\hat{z}_{T}]$ is given by
\[
    \mathbb{E}[\hat{z}_T] = \sum_{t=0}^{T-1} q_{T}(t) z_{t},
\]
i.e.\ in expectation the model reduces to an autoregressive (AR) model with weights $q_T(\cdot)$, and can thus be seen as a non-parametric probabilistic extension of the AR framework.

NPTS uses the following sampling distribution $q_T$, which uses weights that decay exponentially according to the distance of the past observations,
\begin{equation}
        q_T(t) = \frac{\exp(-\alpha \abs{T - t})}{\sum_{t'=0}^{T-1} \exp(-\alpha \abs{T -t})},
\label{eq:expKernel}
\end{equation}
where $\alpha$ is a hyper-parameter that is adjusted based on the data.
In this way, the observations from the recent past are sampled with much higher probability than the observations from the distant past.

The special case, where one uses uniform weights for all the time steps in the training range, i.e., $q_t = 1$, leads to the Climatological forecaster. This can equivalently be achieved by letting $\alpha \rightarrow 0$ in Eq.~\ref{eq:expKernel}. Similarly, by letting $\alpha \rightarrow \infty$, we recover the naive forecaster, which always predicts the last observed value.  One strength of NPTS as a baseline method is that it can be used to smoothly interpolate between these two extremes by varying $\alpha$.

So far we have only described the one step ahead forecast.
To generate forecasts for multiple time steps, one can simply absorb the predictions for the last time steps into the observed targets, and then generate subsequent predictions using the last $T$ targets. For example, prediction for time step $T+t, t > 0$ is generated using the targets $z_t, \ldots, z_{T-1},\hat{z}_T, \ldots, \hat{z}_{T+t-1}$.

\textbf{DeepAR.} \Swist{} contains an auto-regressive RNN time series model, DeepAR, which is similar to
the architectures described in
\citep{flunkert2017deepar,sqf}. DeepAR consists of a RNN
(either using LSTM or GRU cells) that takes the previous time points and co-variates as input. DeepAR then either estimates parameters of a parametric
distribution (see Sec.~\ref{sec:distribution}) or a highly flexible
parameterization of the quantile function. Training and prediction follow the
general approach for auto-regressive models discussed above.

In contrast to the original model description in
\citep{flunkert2017deepar}, DeepAR not only takes the
last target value as an input, but also a number of lagged values that are relevant
for the given frequency. For instance, for hourly data, the lags may be $1$
(previous time point), $1\times 24$ (one day ago), $2\times 24$ (two days
ago), $7\times 24$ (one week ago) etc.

\textbf{Wavenet}~\citep{wavenet} is an auto-regressive neural network with
dilated causal convolutions at its core. In the set of \Swist{} models, it
represents the archetypical auto-regressive \gls{CNN} models. While it was developed
for speech synthesis tasks, it is in essence a time series model that can be
used for time series modeling in other problem domains. In the text-to-speech
application, the output is a bounded signal and in many implementations the
value range is often quantized into discrete bins. This discretized signal is
then modeled using a flexible softmax distribution that can represent arbitrary
distributions over the discrete values, at the cost of discarding ordinal
information.

As discussed in Sec.~\ref{sec:distribution} \Swist{} supports such discretized distributions via a
binned distribution. The neural architecture of Wavenet is not tied to a quantized output,
and can be readily combined with any of the distributions available in \Swist{}.

\section{Experiments}\label{sec:experiments}

In this section we demonstrate the practical effectiveness of a subset of the forecast and anomaly detection models
discussed in Sec.~\ref{sec:models}. 

Table~\ref{tbl:forecasteval} shows the CRPS \citep{sqf} accuracy of different forecast methods in
\Swist{} on the following 11 public datasets:

\begin{itemize}
\item S\&P500: daily difference time series of log-return of stocks from S\&P500
\item electricity: hourly time series of the electricity consumption of 370 customers \citep{Dua:2017}
\item m4: 6 datasets from the M4 competition \citep{makridakisM4concl} with daily, hourly, weekly, monthly, quaterly and yearly frequencies
\item parts: monthly demand for car parts used in \citep{seeger2016}
\item traffic: hourly occupancy rate, between 0 and 1, of 963 car lanes of San Francisco bay area freeways \citep{Dua:2017}
\item wiki10k: daily traffic of 10K Wikipedia pages
\end{itemize}

The hyperparameters of each methods are kept constant across all datasets and we
train with 5000 gradient updates the neural networks models. Each method is run
on a 8 cores machine with 16 GB of RAM. The neural network methods
compare favorably to established methods such as ARIMA or ETS. The total running
time of the methods (training + prediction) is included in Appendix~\ref{sec:running-time}. 
Running times are similar for the \Swist{}-based models. Most of them can be trained and evaluated under an hour. 

Regarding accuracy, there is no overall dominating method. Hence, the experiment illustrates the need 
for a flexible modeling toolkit, such as \Swist{}, that allows to assemble models quickly for the data set and application at hand.

\begin{table}[t]
\tiny{
\makebox[\textwidth]{\begin{tabular}{llllllll}
\toprule
estimator &           \textbf{Auto-ARIMA} &             \textbf{Auto-ETS} &          \textbf{Prophet} &                 \textbf{NPTS} &      \textbf{Transformer} &               \textbf{CNN-QR} &                \textbf{DeepAR} \\
dataset       &                      &                      &                  &                      &                  &                      &                      \\
\midrule
SP500-returns &      0.975 +/- 0.001 &      0.982 +/- 0.001 &  0.985 +/- 0.001 &  \textbf{0.832 +/- 0.000} &  0.836 +/- 0.001 &      0.907 +/- 0.006 &      0.837 +/- 0.002 \\
electricity   &      0.056 +/- 0.000 &      0.067 +/- 0.000 &  0.094 +/- 0.000 &  \textbf{0.055 +/- 0.000} &  0.062 +/- 0.001 &      0.081 +/- 0.002 &      0.065 +/- 0.006 \\
m4-Daily      &      0.024 +/- 0.000 &  \textbf{0.023 +/- 0.000} &  0.090 +/- 0.000 &      0.145 +/- 0.000 &  0.028 +/- 0.000 &      0.026 +/- 0.001 &      0.028 +/- 0.000 \\
m4-Hourly     &      0.040 +/- 0.001 &      0.044 +/- 0.001 &  0.043 +/- 0.000 &      0.048 +/- 0.000 &  0.042 +/- 0.010 &      0.065 +/- 0.008 &  \textbf{0.034 +/- 0.004} \\
m4-Monthly    &  \textbf{0.097 +/- 0.000} &      0.099 +/- 0.000 &  0.132 +/- 0.000 &      0.233 +/- 0.000 &  0.134 +/- 0.002 &      0.126 +/- 0.002 &      0.135 +/- 0.003 \\
m4-Quarterly  &      0.080 +/- 0.000 &  \textbf{0.078 +/- 0.000} &  0.123 +/- 0.000 &      0.255 +/- 0.000 &  0.095 +/- 0.003 &      0.091 +/- 0.000 &      0.091 +/- 0.001 \\
m4-Weekly     &  \textbf{0.050 +/- 0.000} &      0.051 +/- 0.000 &  0.108 +/- 0.000 &      0.296 +/- 0.001 &  0.075 +/- 0.005 &      0.056 +/- 0.000 &      0.072 +/- 0.003 \\
m4-Yearly     &      0.124 +/- 0.000 &      0.126 +/- 0.000 &  0.156 +/- 0.000 &      0.355 +/- 0.000 &  0.127 +/- 0.004 &      0.121 +/- 0.000 &  \textbf{0.120 +/- 0.002} \\
parts         &      1.403 +/- 0.002 &      1.342 +/- 0.002 &  1.637 +/- 0.002 &      1.355 +/- 0.002 &  1.000 +/- 0.003 &  \textbf{0.901 +/- 0.000} &      0.972 +/- 0.005 \\
traffic       &                 - &      0.462 +/- 0.000 &  0.273 +/- 0.000 &      0.162 +/- 0.000 &  0.132 +/- 0.010 &      0.186 +/- 0.002 &  \textbf{0.127 +/- 0.004} \\
wiki10k       &      0.610 +/- 0.001 &      0.841 +/- 0.001 &  0.681 +/- 0.000 &      0.452 +/- 0.000 &  0.294 +/- 0.008 &      0.314 +/- 0.002 &  \textbf{0.292 +/- 0.021} \\
\bottomrule
\end{tabular}}
}

\caption{CRPS error for all methods. Ten runs are done for each method the Mean and std are computed over 10 runs.
  Missing value indicates the method did not complete in 24~hours.}
\label{tbl:forecasteval}
\end{table}

\begin{figure}[tb]
	\centering
	\includegraphics{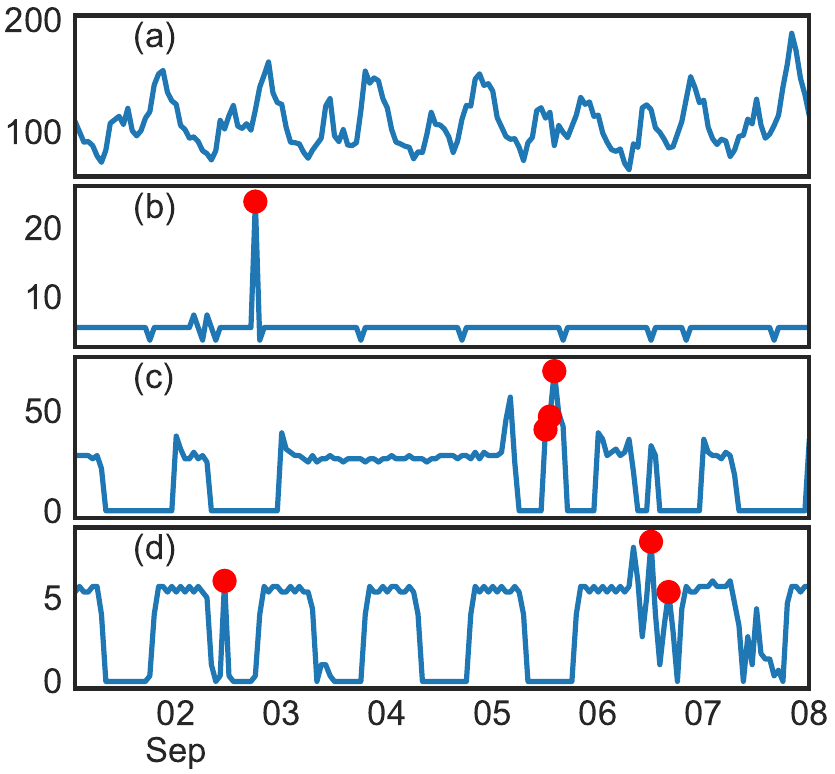}
	\caption{Examples of anomalies detected using a trained DeepAR forecast model
    on the electricity dataset.}
	\label{fig:anomaly}
\end{figure}

The accuracy of neural forecast methods for time series anomaly detection has
been previously demonstrated~\citep{shipmonTimeSeriesAnomaly2017}. Here, we
provide exemplary plots for anomalies detected using the forecast models
discussed above. Starting from a trained DeepAR forecast model on the electricity 
dataset, we use the CDF at each time point and
mark values as anomalies using a $p$-value of $10^{-4}$. The results are
depicted in Fig.~\ref{fig:anomaly}. The data consists of hourly observations and
we plot a detection window of one week. Most of the time series do not
exhibit anomalies in this time frame, as shown in panel (a). Panel (b)--(d)
depict example time series for which the model detected anomalous behavior. The
detected anomalies are
points that do not match the series historical behavior such as seasonality or noise
level. These plots demonstrate the applicability of the \Swist{} neural time series models for anomaly
detection. Clearly, further research and experiments are necessary to e.g.\ quantify
the relation between forecast accuracy and detection accuracy.

\section{Related Work}\label{sec:rel_work}

Deep learning frameworks, such as~\citep{chen2015mxnet,paszke2017automatic,tensorflow} are growing in popularity.
In recent years, more application-specific toolkits have appeared primarily in those areas where deep learning has been overwhelmingly successful, such as computer vision and language-related tasks, but also increasingly beyond~\citep{sockeye,mxfusion,bingham2018pyro}. For forecasting, we are not aware of packages based on modern deep learning toolkits. 

Given the practical importance of time series modeling and forecasting, a number
of commercial and open-source toolkits exist. The R-forecast
package~\citep{hyndman2008R} and other
packages such as~\citep{prophet17,Scott2014PredictingTP} provide a plethora of models and
tooling for classical forecasting methods. To the best of our knowledge,
\Swist{} is the first toolkit for time series modeling based on a modern deep
learning framework. Some forecasting packages in R contain neural forecasting
models\footnote{For example
\url{https://cran.r-project.org/web/packages/nnfor/nnfor.pdf}}, however these
pre-date modern deep learning methods and only contain stand-alone
implementations of simple local models, in particular, they lack state-of-the-art
architectures. As deep learning based models are
gaining popularity in
forecasting~\citep{flunkert2017deepar,sqf,kari2017,laptev2017},
\Swist{} offers abstractions to implement many of the currently proposed models.
\Swist{} contains pure deep learning based models, and components from classical
time series models such as Kalman filters~\citep{seeger2017autodifferentiating}.
A recent trend in the machine learning literature are fusions of deep and
probabilistic models~\citep{makridakisM4concl,rangapuram2018,smyl}. By having
probabilistic building blocks available, \Swist{} allows for the systematic
exploration of these models.

Most forecasting libraries that we are aware of are primarily geared towards running pre-built models and less so towards the scientific use case of model development.
\Swist{} differs from this by providing all the components and tools necessary
for rapid model prototyping, development and benchmarking against pre-assembled models.
It tries to strike a balance between a short path to production and rapid scientific exploration.
The experimentation process is not as heavy-weight as in typical production forecasting systems~\citep{bose2017probabilistic}, because the library has native support for local, stand-alone execution.

\section{Conclusion}\label{sec:outro}

We introduced \Swist{}, a toolkit for building
time series models based on deep learning and
probabilistic modeling techniques. 
By offering tooling and abstractions such as 
probabilistic models, basic neural building blocks, 
human-readable model logging for increased reproducability and unified I/O \& evaluation, 
 \Swist{} allows scientists to rapidly develop new
time series models for common tasks such as forecasting or anomaly detection. 
\Swist{} is actively used at Amazon in a variety of internal 
and external use-cases (including production) where it has helped scientists 
to address time series modelling challenges.

\Swist{}'s pre-bundled implementations of state-of-the-art models allow
easy benchmarking of new algorithms. We demonstrated this in a large
scale experiment of running the pre-bundled models on different datasets and
comparing their accuracy with classical approaches. Such experiments are a 
 first step towards a more thorough understanding of neural architectures for 
time series modelling. More and more fine-grained 
experiments such as ablation studies and experiments with controlled data are needed as next steps. 
\Swist{} provides the tooling necessary for such future work.

\bibliography{ref.bib}

\begin{thebibliography}{46}
\providecommand{\natexlab}[1]{#1}
\providecommand{\url}[1]{\texttt{#1}}
\expandafter\ifx\csname urlstyle\endcsname\relax
  \providecommand{\doi}[1]{doi: #1}\else
  \providecommand{\doi}{doi: \begingroup \urlstyle{rm}\Url}\fi

\bibitem[Abadi et~al.(2016)]{tensorflow}
Mart\'{\i}n Abadi et~al.
\newblock Tensorflow: A system for large-scale machine learning.
\newblock In \emph{Proceedings of the 12th USENIX Conference on Operating
  Systems Design and Implementation}, OSDI'16, pages 265--283, Berkeley, CA,
  USA, 2016. USENIX Association.
\newblock ISBN 978-1-931971-33-1.

\bibitem[Biessmann et~al.(2018)Biessmann, Salinas, Schelter, Schmidt, and
  Lange]{missingValue}
Felix Biessmann, David Salinas, Sebastian Schelter, Philipp Schmidt, and Dustin
  Lange.
\newblock {"Deep"} learning for missing value imputation in tables with
  non-numerical data.
\newblock In \emph{Proceedings of the 27th ACM International Conference on
  Information and Knowledge Management}, CIKM '18, pages 2017--2025, New York,
  NY, USA, 2018. ACM.
\newblock ISBN 978-1-4503-6014-2.

\bibitem[Bingham et~al.(2018)]{bingham2018pyro}
Eli Bingham et~al.
\newblock {Pyro: Deep Universal Probabilistic Programming}.
\newblock \emph{Journal of Machine Learning Research}, 2018.

\bibitem[B{\"o}se et~al.(2017)]{bose2017probabilistic}
Joos-Hendrik B{\"o}se et~al.
\newblock Probabilistic demand forecasting at scale.
\newblock \emph{PVLDB}, 10\penalty0 (12):\penalty0 1694--1705, 2017.

\bibitem[Box and Cox(1964)]{boxcox1964}
G.~E.~P. Box and D.~R. Cox.
\newblock An analysis of transformations.
\newblock \emph{Journal of the Royal Statistical Society. Series B
  (Methodological)}, 26\penalty0 (2):\penalty0 211--252, 1964.

\bibitem[Chapados(2014)]{chapados2014}
Nicolas Chapados.
\newblock Effective {B}ayesian modeling of groups of related count time series.
\newblock In \emph{Proceedings of The 31st International Conference on Machine
  Learning}, pages 1395--1403, 2014.

\bibitem[Chen et~al.(2015)]{chen2015mxnet}
Tianqi Chen et~al.
\newblock Mxnet: A flexible and efficient machine learning library for
  heterogeneous distributed systems.
\newblock \emph{arXiv preprint arXiv:1512.01274}, 2015.

\bibitem[Dai et~al.(2018)Dai, Meissner, and Lawrence]{mxfusion}
Zhenwen Dai, Eric Meissner, and Neil~D. Lawrence.
\newblock {MXFusion}: A modular deep probabilistic programming library.
\newblock In \emph{NIPS Workshop MLOSS (Machine Learning Open Source
  Software)}, 2018.

\bibitem[Dheeru and Karra~Taniskidou(2017)]{Dua:2017}
Dua Dheeru and Efi Karra~Taniskidou.
\newblock {UCI} machine learning repository, 2017.
\newblock URL \url{http://archive.ics.uci.edu/ml}.

\bibitem[Dillon et~al.(2017)]{tensorflow_distributions}
Joshua~V. Dillon et~al.
\newblock Tensorflow distributions.
\newblock \emph{CoRR}, abs/1711.10604, 2017.

\bibitem[Durbin and Koopman(2012)]{durbin2012time}
James Durbin and Siem~Jan Koopman.
\newblock \emph{Time series analysis by state space methods}, volume~38.
\newblock OUP Oxford, 2012.

\bibitem[Faloutsos et~al.(2018)Faloutsos, Gasthaus, Januschowski, and
  Wang]{Faloutsos2018}
Christos Faloutsos, Jan Gasthaus, Tim Januschowski, and Yuyang Wang.
\newblock Forecasting big time series: old and new.
\newblock \emph{Proceedings of the VLDB Endowment}, 11\penalty0 (12):\penalty0
  2102--2105, 2018.

\bibitem[Flunkert et~al.(to appear)Flunkert, Salinas, Gasthaus, and
  Januschowski]{flunkert2017deepar}
Valentin Flunkert, David Salinas, Jan Gasthaus, and Tim Januschowski.
\newblock {DeepAR}: Probabilistic forecasting with autoregressive recurrent
  networks.
\newblock \emph{International Journal of Forecasting}, to appear.

\bibitem[Fraccaro et~al.(2016)Fraccaro, S{\o}nderby, Paquet, and
  Winther]{fraccaro2016sequential}
Marco Fraccaro, S{\o}ren~Kaae S{\o}nderby, Ulrich Paquet, and Ole Winther.
\newblock Sequential neural models with stochastic layers.
\newblock In \emph{Advances in neural information processing systems}, pages
  2199--2207, 2016.

\bibitem[Gasthaus(2016)]{gasthaus2016}
Jan Gasthaus.
\newblock Non-parametric time series forecaster.
\newblock Technical report, Amazon, 2016.

\bibitem[Gasthaus et~al.(2019)Gasthaus, Benidis, Wang, Rangapuram, Salinas,
  Flunkert, and Januschowski]{sqf}
Jan Gasthaus, Konstantinos Benidis, Yuyang Wang, Syama~Sundar Rangapuram, David
  Salinas, Valentin Flunkert, and Tim Januschowski.
\newblock Probabilistic forecasting with {Spline Quantile Function RNN}s.
\newblock In Kamalika Chaudhuri and Masashi Sugiyama, editors,
  \emph{Proceedings of Machine Learning Research}, volume~89 of
  \emph{Proceedings of Machine Learning Research}, pages 1901--1910. PMLR,
  16--18 Apr 2019.
\newblock URL \url{http://proceedings.mlr.press/v89/gasthaus19a.html}.

\bibitem[Girard et~al.(2003)Girard, Rasmussen, Candela, and
  Murray-Smith]{girard2003gaussian}
Agathe Girard, Carl~Edward Rasmussen, Joaquin~Quinonero Candela, and Roderick
  Murray-Smith.
\newblock Gaussian process priors with uncertain inputs application to
  multiple-step ahead time series forecasting.
\newblock In \emph{Advances in neural information processing systems}, pages
  545--552, 2003.

\bibitem[Hieber et~al.(2018)]{sockeye}
Felix Hieber et~al.
\newblock The {Sockeye} neural machine translation toolkit at {AMTA} 2018.
\newblock In \emph{{AMTA} {(1)}}, pages 200--207. Association for Machine
  Translation in the Americas, 2018.

\bibitem[Hunter(2007)]{matplotlib}
J.~D. Hunter.
\newblock Matplotlib: A {2D} graphics environment.
\newblock \emph{Computing In Science \& Engineering}, 9\penalty0 (3):\penalty0
  90--95, 2007.

\bibitem[Hyndman et~al.(2008)Hyndman, Koehler, Ord, and Snyder]{hyndman2008}
R.~Hyndman, A.~B. Koehler, J.~K. Ord, and R.~D. Snyder.
\newblock \emph{Forecasting with Exponential Smoothing: The State Space
  Approach}.
\newblock Springer Series in Statistics. Springer, 2008.
\newblock ISBN 9783540719182.

\bibitem[Hyndman and Athanasopoulos(2017)]{hyndman2017forecasting}
Rob~J Hyndman and George Athanasopoulos.
\newblock Forecasting: Principles and practice.
\newblock \emph{www. otexts. org/fpp.}, 987507109, 2017.

\bibitem[Hyndman and Khandakar(2008)]{hyndman2008R}
Rob~J Hyndman and Yeasmin Khandakar.
\newblock Automatic time series forecasting: the forecast package for {R}.
\newblock \emph{Journal of Statistical Software}, 2008.

\bibitem[Koenker(2005)]{koenker_2005}
Roger Koenker.
\newblock \emph{Quantile Regression}.
\newblock Econometric Society Monographs. Cambridge University Press, 2005.

\bibitem[Krishnan et~al.(2017)Krishnan, Shalit, and
  Sontag]{krishnan2017structured}
Rahul~G Krishnan, Uri Shalit, and David Sontag.
\newblock Structured inference networks for nonlinear state space models.
\newblock In \emph{AAAI}, pages 2101--2109, 2017.

\bibitem[Laptev et~al.(2017)Laptev, Yosinsk, Li~Erran, and Smyl]{laptev2017}
Nikolay Laptev, Jason Yosinsk, Li~Li~Erran, and Slawek Smyl.
\newblock Time-series extreme event forecasting with neural networks at {Uber}.
\newblock In \emph{ICML Time Series Workshop}. 2017.

\bibitem[Maddix et~al.(2018)Maddix, Wang, and Smola]{maddix2018deep}
Danielle~C Maddix, Yuyang Wang, and Alex Smola.
\newblock Deep factors with {G}aussian processes for forecasting.
\newblock \emph{arXiv preprint arXiv:1812.00098}, 2018.

\bibitem[Makridakis et~al.(2018)Makridakis, Spiliotis, and
  Assimakopoulos]{makridakisM4concl}
Spyros Makridakis, Evangelos Spiliotis, and Vassilios Assimakopoulos.
\newblock The {M4} competition: Results, findings, conclusion and way forward.
\newblock \emph{International Journal of Forecasting}, 34\penalty0
  (4):\penalty0 802 -- 808, 2018.
\newblock ISSN 0169-2070.
\newblock \doi{https://doi.org/10.1016/j.ijforecast.2018.06.001}.

\bibitem[Meng et~al.(2015)]{sparkml}
Xiangrui Meng et~al.
\newblock Mllib: Machine learning in apache spark.
\newblock \emph{CoRR}, abs/1505.06807, 2015.
\newblock URL \url{http://arxiv.org/abs/1505.06807}.

\bibitem[Ng and Jordan(2002)]{ng02}
Andrew~Y. Ng and Michael~I. Jordan.
\newblock On discriminative vs. generative classifiers: A comparison of
  logistic regression and naive {B}ayes.
\newblock In T.~G. Dietterich, S.~Becker, and Z.~Ghahramani, editors,
  \emph{Advances in Neural Information Processing Systems 14}, pages 841--848.
  MIT Press, 2002.

\bibitem[Paszke et~al.(2017)]{paszke2017automatic}
Adam Paszke et~al.
\newblock Automatic differentiation in {PyTorch}.
\newblock In \emph{NIPS-W}, 2017.

\bibitem[Pedregosa et~al.(2011)]{scikitlearn}
Fabian Pedregosa et~al.
\newblock Scikit-learn: Machine learning in {P}ython.
\newblock \emph{J. Mach. Learn. Res.}, 12:\penalty0 2825--2830, nov 2011.
\newblock ISSN 1532-4435.

\bibitem[Rangapuram et~al.(2018)Rangapuram, Seeger, Gasthaus, Stella, Wang, and
  Januschowski]{rangapuram2018}
Syama~Sundar Rangapuram, Matthias Seeger, Jan Gasthaus, Lorenzo Stella, Yuyang
  Wang, and Tim Januschowski.
\newblock Deep state space models for time series forecasting.
\newblock In \emph{Advances in Neural Information Processing Systems}, 2018.

\bibitem[Rezende and Mohamed(2015)]{rezendeVariationalInferenceNormalizing2015}
Danilo~Jimenez Rezende and Shakir Mohamed.
\newblock Variational inference with normalizing flows.
\newblock In \emph{Proceedings of the 32Nd International Conference on
  International Conference on Machine Learning - Volume 37}, ICML'15, pages
  1530--1538. JMLR.org, 2015.

\bibitem[Scott and Varian(2014)]{Scott2014PredictingTP}
Steven~L. Scott and Hal~R. Varian.
\newblock Predicting the present with {B}ayesian structural time series.
\newblock \emph{IJMNO}, 5:\penalty0 4--23, 2014.

\bibitem[Seeger et~al.(2017)Seeger, Hetzel, Dai, Meissner, and
  Lawrence]{seeger2017autodifferentiating}
Matthias Seeger, Asmus Hetzel, Zhenwen Dai, Eric Meissner, and Neil~D.
  Lawrence.
\newblock Auto-differentiating linear algebra, 2017.

\bibitem[Seeger et~al.(2016)Seeger, Salinas, and Flunkert]{seeger2016}
Matthias~W Seeger, David Salinas, and Valentin Flunkert.
\newblock Bayesian intermittent demand forecasting for large inventories.
\newblock In \emph{Advances in Neural Information Processing Systems}, pages
  4646--4654, 2016.

\bibitem[Shipmon et~al.(2017)Shipmon, Gurevitch, Piselli, and
  Edwards]{shipmonTimeSeriesAnomaly2017}
Dominique~T Shipmon, Jason~M Gurevitch, Paolo~M Piselli, and Steve Edwards.
\newblock Time {{Series Anomaly Detection}}.
\newblock \emph{arXiv:1708.03665 [stat.ML]}, page~9, 2017.

\bibitem[Smyl et~al.(2018)Smyl, Ranganathan, and Pasqua]{smyl}
Slawek Smyl, Jai Ranganathan, and Andrea Pasqua.
\newblock {M4} forecasting competition: Introducing a new hybrid {ES-RNN}
  model.
\newblock \url{https://eng.uber.com/m4-forecasting-competition/}, 2018.

\bibitem[Sutskever et~al.(2014)Sutskever, Vinyals, and Le]{sutskever2014}
Ilya Sutskever, Oriol Vinyals, and Quoc~V Le.
\newblock Sequence to sequence learning with neural networks.
\newblock In \emph{Advances in Neural Information Processing Systems}, pages
  3104--3112, 2014.

\bibitem[Taylor and Letham(2017)]{prophet17}
Sean~J Taylor and Benjamin Letham.
\newblock Forecasting at scale.
\newblock \emph{PeerJ Preprints}, 5:\penalty0 e3190v2, September 2017.
\newblock ISSN 2167-9843.
\newblock \doi{10.7287/peerj.preprints.3190v2}.

\bibitem[van~den Oord et~al.(2016)]{wavenet}
A{\"{a}}ron van~den Oord et~al.
\newblock {WaveNet}: {A} generative model for raw audio.
\newblock \emph{CoRR}, abs/1609.03499, 2016.

\bibitem[Vaswani et~al.(2017)]{attention}
Ashish Vaswani et~al.
\newblock Attention is all you need.
\newblock In I.~Guyon, U.~V. Luxburg, S.~Bengio, H.~Wallach, R.~Fergus,
  S.~Vishwanathan, and R.~Garnett, editors, \emph{Advances in Neural
  Information Processing Systems 30}, pages 5998--6008. Curran Associates,
  Inc., 2017.

\bibitem[Wang et~al.(2019)Wang, Smola, Maddix, Gasthaus, Foster, and
  Januschowski]{wang2019deep}
Yuyang Wang, Alex Smola, Danielle Maddix, Jan Gasthaus, Dean Foster, and Tim
  Januschowski.
\newblock Deep factors for forecasting.
\newblock In \emph{International Conference on Machine Learning}, pages
  6607--6617, 2019.

\bibitem[Wen et~al.(2017)Wen, Torkkola, and Narayanaswamy]{kari2017}
Ruofeng~Wen Wen, Kari Torkkola, and Balakrishnan Narayanaswamy.
\newblock A multi-horizon quantile recurrent forecaster.
\newblock In \emph{NIPS Time Series Workshop}. 2017.

\bibitem[Xu et~al.(2016)Xu, Liu, Jiang, and Yu]{Xu2016quantile}
Qifa Xu, Xi~Liu, Cuixia Jiang, and Keming Yu.
\newblock Quantile autoregression neural network model with applications to
  evaluating value at risk.
\newblock \emph{Applied Soft Computing}, 49:\penalty0 1--12, 2016.

\bibitem[Zhang et~al.(1998)Zhang, Patuwo, and Hu]{zhang1998forecasting}
Guoqiang Zhang, B~Eddy Patuwo, and Michael~Y Hu.
\newblock Forecasting with artificial neural networks:: The state of the art.
\newblock \emph{International journal of forecasting}, 14\penalty0
  (1):\penalty0 35--62, 1998.

\end{thebibliography}

\clearpage
\printglossary[type=\acronymtype]

\appendix
\clearpage

\section{Experiment details and running times}
\label{sec:running-time}

Here, we provide more details about the experiments in Sec.~\ref{sec:experiments} Table~\ref{tbl:forecasteval}.
All neural network models were trained with batch size $32$,  using $5000$
overall batches (gradient updates).  We used the ADAM optimizer with an initial
learning rate of $10^{-3}$. The learning rate was halved after $300$ batches if
there was no reduction in training loss. The gradient norm was clipped at a
magnitude of 10. For DeepAR and Transformer a student's t distribution was used
for the one-step ahead prediction, which works well with noisy data. For models
that generate sample paths (all except CNN-QR), 100 sample paths were drawn for
the evaluation. For CNN-QR the quantiles $0.1, 0.2, \dots, 0.9$ were estimated.
These quantiles were also used during the evaluation of CRPS for all methods.

The below table gives more detailed information about the datasets and
the evaluation scenario used. In the table freq is the granularity of the dataset, prediction
length is the forecast horizon that was used for the evaluation. Rolling
evaluation indicates whether the evaluation was done over multiple forecasts.
For ``--'' the evaluation was done for a single forecast on the last time window
in each time series. In the case of rolling evaluations, a number of consecutive
forecasts were evaluated shifted, by the prediction length. For instance, for electricity
the last 7 days were used for evaluation where a forecast was generated for each day.
\begin{center}
  \small
\begin{tabular}{llrr}
\toprule
       dataset & freq           &  prediction length & rolling window evaluation \\
\midrule
    electricity & hourly        &           24       &  7 \\
       m4-Daily & daily         &           14       &  - \\
      m4-Hourly & hourly        &           48       &  - \\
     m4-Monthly & monthly       &           18       &  - \\
   m4-Quarterly & quarterly     &            8       &  - \\
      m4-Weekly & weekly        &           13       &  - \\
      m4-Yearly & yearly        &            6       &  - \\
          parts & monthly       &            8       &  - \\
  SP500-returns & business day  &           30       &  5  \\
        traffic & hourly        &           24       &  7 \\
        wiki10k & daily         &           60       &  - \\
\bottomrule
\end{tabular}
\end{center}

\begin{figure}[h]
  \centering
  \includegraphics[width=0.4\columnwidth]{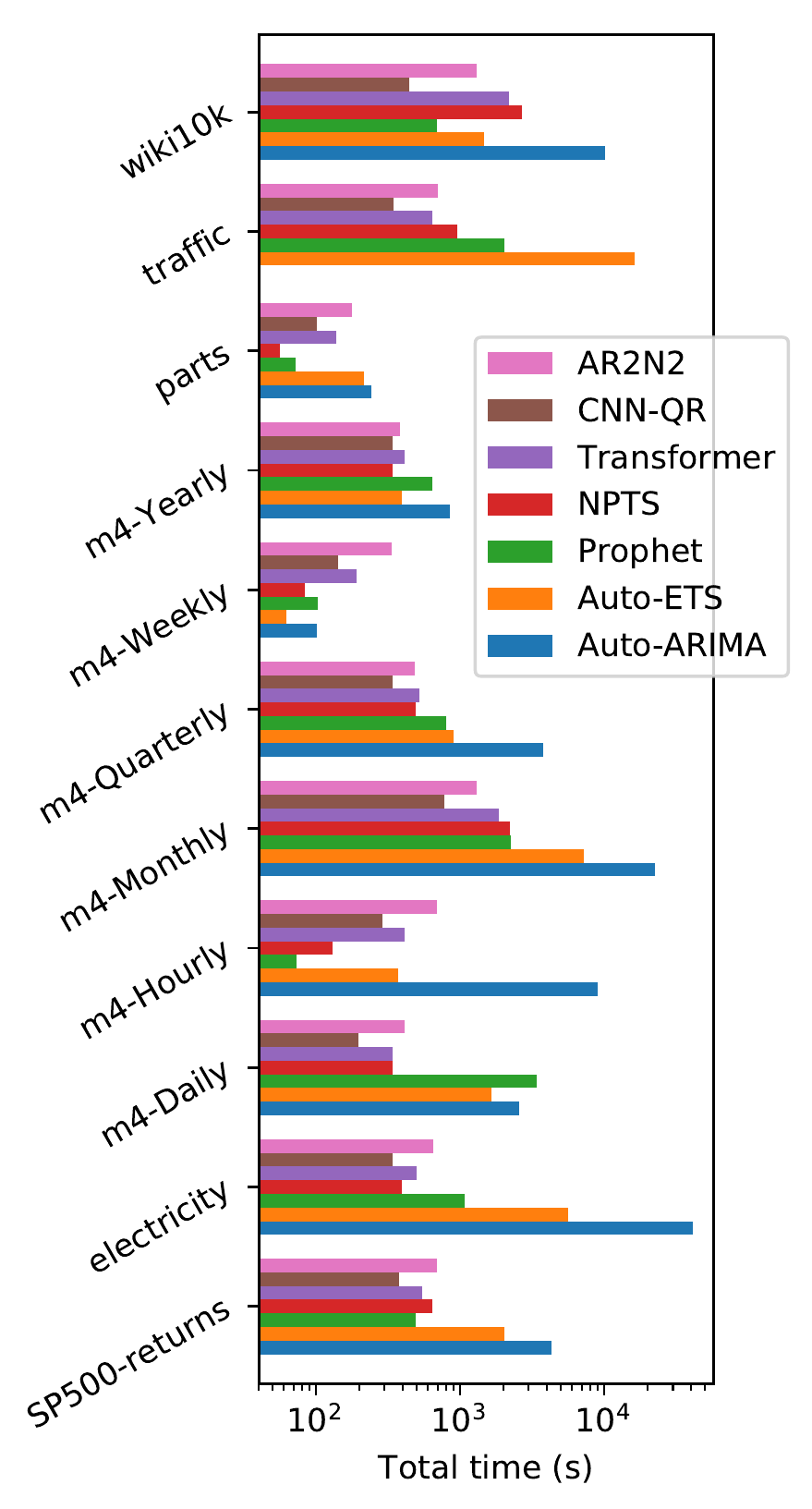}
  \caption{Running time in second for all datasets and methods (logarithmic scale). DeepAR is labeled as AR2N2 in this figure.}
 \label{fig:running_time}
\end{figure}

Figure~\ref{fig:running_time} shows the running times for the experiments in Sec.~\ref{sec:experiments} Table~\ref{tbl:forecasteval}.

\section{Reproducibility of experiments in \Swist{}}
\label{sec:reproducibility}

To support scientists in their work, \Swist{} supports tracking the entire
configuration of an experiment. This is achieved through a Python
decorator \lstinline|@validated| that tracks all constructor arguments of models
and components. The below listing \ref{sec:newmodel} demonstrates how to create
a new neural network based model in \Swist{} (see Sec.~\ref{sec:modelbuilding}
for a discussion of the necessary components). The new estimator (\lstinline|MyEstimator|)
uses the \lstinline|@validated| decorator and Python3 type annotations for the
constructor arguments. \Swist{} can serialize any class that is annotated in
this way to json or to a human readable format. The \lstinline|MyEstimator| has a 
\lstinline|trainer| argument with a default value. This \lstinline|Trainer| class 
contains parameters potentially with default values itself.

For example, if we instantiate this new estimator in an experiment
\begin{lstlisting}[language=python,breaklines=true,emph={MyEstimator}]
estim = MyEstimator(
  freq='D', context_length=50, prediction_length=20
)
\end{lstlisting}
and then log the instance, the following text output is generated:
\begin{lstlisting}[basicstyle=\ttm,stringstyle=\ttm,emphstyle=\ttm]
MyEstimator(
  freq='D',
  context_length=50,
  prediction_length=20,
  act_type='relu',
  cells=[40, 40, 40],
  trainer=gluonts.trainer.Trainer(
    batch_size=32,
    clip_gradient=10.0,
    epochs=50,
    learning_rate=0.001,
    learning_rate_decay_factor=0.5,
    minimum_learning_rate=5e-05,
    patience=10,
    weight_decay=1e-08
  )
)
\end{lstlisting}
All arguments including default values and the configuration of all nested
components are logged. The resulting string is not only readable for the user,
it also allows the user to copy paste the code to instantiate the exact same
model configuration. During a backtest \Swist{} logs the entire configuration of the
train/test split, the model and e.g.\ the evaluation configuration in this way, allowing the user to inspect all parameter settings in hindsight and to fully re-create the experimental setup.

\section{Sample model code}
\label{sec:newmodel}

Here, we provide a full example of a new time series model using 
the \Swist{} abstractions, extending Sec.~\ref{sec:modelbuilding}. Such 
a model can be integrated into the workflow at the beginning of 
Sec.~\ref{sec:library} by replacing \lstinline|DeepAREstimator| in line~\ref{lst:estimator} in 
Listing~\ref{lst:workflow} with \lstinline|MyEstimator|.

\begin{lstlisting}[language=python,style=numbers,breaklines=true,emph={DatasetRepository,DeepAREstimator,Evaluator,Trainer,MyTrainNetwork,MyPredNetwork,GluonEstimator,HybridBlock,QuantileForecast,InstanceSplitter, Predictor, ExpectedNumInstanceSampler, FieldName, Transformation, RepresentableBlockPredictor,__init__}]
from typing import List
from mxnet import gluon
from gluonts.model.estimator import GluonEstimator 
from gluonts.model.predictor import Predictor, RepresentableBlockPredictor
from gluonts.trainer import Trainer
from gluonts.transform import InstanceSplitter, FieldName, Transformation, ExpectedNumInstanceSampler
from gluonts.core.component import validated
from gluonts.support.util import copy_parameters

class MyTrainNetwork(gluon.HybridBlock):
  def __init__(self, prediction_length, cells, act_type, **kwargs):
    super().__init__(**kwargs)
    self.prediction_length = prediction_length
  
    with self.name_scope():
      # Set up a network that predicts the target
      self.nn = gluon.nn.HybridSequential()
      for c in cells:
        self.nn.add(gluon.nn.Dense(
          units=c, activation=act_type
        ))
      self.nn.add(gluon.nn.Dense(
        units=self.prediction_length,
        activation=act_type
      ))

  def hybrid_forward(self, F, past_target, future_target):
    prediction = self.nn(past_target)
    # calculate L1 loss to learn the median
    return (prediction - future_target).abs() \
      .mean(axis=-1)

class MyPredNetwork(MyTrainNetwork):
  # The prediction network only receives 
  # past_target and returns predictions
  def hybrid_forward(self, F, past_target):
    prediction = self.nn(past_target)
    return prediction.expand_dims(axis=1)

class MyEstimator(GluonEstimator):
  @validated()
  def __init__(
    self,
    freq: str,
    prediction_length: int,
    act_type: str = "relu",
    context_length: int = 30,
    cells: List[int] = [40, 40, 40], 
    trainer: Trainer = Trainer(epochs=10)
  ) -> None:
    super().__init__(trainer=trainer)
    self.freq = freq
    self.prediction_length = prediction_length
    self.act_type = act_type
    self.context_length = context_length 
    self.cells = cells
 	
    def create_transformation(self):
	# Model specific input transform
	# Here we use a transformation that randomly 
	# selects training samples from all series.
	return InstanceSplitter(target_field=FieldName.TARGET,
	is_pad_field=FieldName.IS_PAD,
	start_field=FieldName.START,
	forecast_start_field=FieldName.FORECAST_START,
	train_sampler=ExpectedNumInstanceSampler(num_instances=1),
	past_length=self.context_length,
	future_length=self.prediction_length,
	)


    def create_training_network(self) -> MyTrainNetwork:
	return MyTrainNetwork(
	prediction_length=self.prediction_length, 
	cells=self.cells,
	act_type=self.act_type,
	)
 	
 	def create_predictor(self, 
 	transformation: Transformation, 
 	trained_network: gluon.HybridBlock
  	) -> Predictor:
 	prediction_network = MyPredNetwork(
 	prediction_length=self.prediction_length, 
 	cells=self.cells,
 	act_type=self.act_type,
 	)
 	
 	copy_parameters(trained_network, prediction_network)
 	
 	return RepresentableBlockPredictor(
 	input_transform=transformation,
 	prediction_net=prediction_network,
 	batch_size=self.trainer.batch_size,
 	freq=self.freq,
 	prediction_length=self.prediction_length,
 	ctx=self.trainer.ctx,
 	)
\end{lstlisting}

\end{document}